# Fusion of Deep and Shallow Features for Face Kinship Verification

1st Belabbaci El Ouanas
*LIMED Laboratory*
*Faculty of Technologie , University of*
Bejaia, 06000 Bejaia, Algeria
elouanas.belabbaci@univ-bejaia.dz

2nd Khammari Mohammed
*LIMED Laboratory*
*Faculty of Exact Sciences, University of*
Bejaia, 06000 Bejaia, Algeria
mohammed.khammari@univ-bejaia.dz

3rd Chouchane Ammar
*dept. Electrical Engineering*
*LI3C, University of Mohamed Khider*
Biskra, Algeria
chouchane_ammar@yahoo.com

4th Mohcene Bessaoudi
*dept. Electrical Engineering*
*LI3C, University of Mohamed Khider*
Biskra, Algeria
bessaoudi.mohcene@gmail.com

5th Abdelmalik Ouamane
*dept. Electrical Engineering*
*LI3C, University of Mohamed Khider*
Biskra, Algeria
ouamaneabdealmalik@yahoo.fr

6th Akram Abderraouf Gharbi
*dept. Electrical Engineering*
*University Mohamed Khider of Biskra*
Biskra, Algeria
akram.gharbi@univ-biskra.dz

*Abstract*—Kinship verification from face images is a novel and formidable challenge in the realms of pattern recognition and computer vision. This work makes notable contributions by incorporating a preprocessing technique known as Multiscale Retinex (MSR), which enhances image quality. Our approach harnesses the strength of complementary deep (VGG16) and shallow texture descriptors (BSIF) by combining them at the score level using Logistic Regression (LR) technique. We assess the effectiveness of our approach by conducting comprehensive experiments on three challenging kinship datasets: Cornell Kin Face, UB Kin Face and TS Kin Face

## I. INTRODUCTION

The objective of kinship verification from face images is to ascertain the biological relationship between two in- dividuals by analyzing their faces appearances [1]. Recognizing kinship is an extremely demanding task due to the complexity of facial features conveyed by the face, including identity, age, gender, ethnicity, and expression [1]. Furthermore, the identification of kinship is a significant area of research with a wide range of applications. It can be utilized for organizing photo albums, constructing family trees, aiding in forensic investigations, annotating images, and even assisting in the identification of missing or wanted individuals [2]. While DNA is traditionally used for kinship verification, an automated algorithm that utilizes face images can provide cost-effective and real-time results for kinship verification. Previous research on kinship verification [3], [4], [5] has predominantly focused on utilizing similar facial features. Recent advancements in machine learning have demonstrated that the using of deep features [3], [6], [5], [7] can yield optimal performance in kinship verification. Combining multiple methods is often beneficial in achieving the highest level of performance.

Multilinear Subspace Learning (MSL) is indeed a powerful machine learning technique that facilitates the identification of discriminant features from various feature extraction approaches, each operating at different scales. It allows for the identification of hidden patterns within large datasets and can be utilized to unveil relationships between different variables. The utilization of MSL with tensor data has emerged as a robust method for kinship verification tasks. Among the most important algorithms utilized to enhance kinship verification, [8] introduced Multilinear Side-Information based Discriminant Analysis (MSIDA). MSIDA typically performs the projection of the input region tensor into a novel multilinear subspace, which results in an increased distance between samples belonging to different classes and a decreased distance among samples belonging to the same class. Another intriguing algorithm in this context is Tensor Cross-View Quadratic Analysis (TXQDA) [9]. TXQDA not only preserves the inherent data structure and enhances the interval between samples but also effectively addresses the challenge of limited sample sizes while simultaneously reducing computational expenses.

In our study, we aim to harness these aspects to present a robust and reliable face kinship verification system that can effectively address the challenges discussed earlier. To achieve this goal, we make several contributions and perform extensive testing on three standard datasets, namely Cornell Kin Face,UB Kin Face and TS Kin Face. The primary contributions of our work can be outlined as follows:
- Utilizing of an effective preprocessing technique called Multiscale Retinex (MSR) to enhance the restoration of image color and improve overall image quality.

Identify applicable funding agency here. If none, delete this.

- Subspace projection and dimensionality reduction are accomplished by employing the effective TXQDA algorithm, which mainly depends multidimensional data representation.
- To enhance our feature extraction process, we incorporate score level fusion utilizing Logistic Regression (LR). Our fusion technique combines the shallow texture features BSIF with deep features extracted from the VGG16 model, exploiting their complementary nature to improved the performance in kinship verification.

The remainder of the essay is structured as follows: Section 2 presents our methodology through introducing the preprocessing-based (MSR), as well as the incorporation of deep and shallow texture features. In Section 3, we provide details about experimental setup which are employed in our study and discuss the results obtained from the experiments. Finally, conclusion are given in Section 4

## II. METHODOLOGY

In this section, we present the design of the face kinship verification system proposed in our work. As illustrated in Fig.1. The framework comprises four key components: (A) Face pre-processing, (B) Feature extraction, (C) Multilinear subspace learning and (D) Matching and fusion. We provide a comprehensive discussion of each stage in the following sections.

### A. Pre-processing

Preprocessing plays a crucial role in face kinship verification. The primary purpose of preprocessing is to enhance the quality and reliability of the input data before it undergoes further analysis and comparison. In this work, MTCNN [10] method is utilized to detect the facial region in the images. Subsequently, we used the MSR algorithm [11] as an image enhancement technique. Figure 2 demonstrates an example of test images processing using MSR. The MSR algorithm enhances the quality of images by expanding their dynamic range and maintaining color accuracy.

### B. Features extraction

Feature extraction is a critical step in face kinship verification [2]. In this study, we employ the VGG16 model to extract deep features from facial images. VGG16 short for Visual Geometry Group 16 layers, is a pretrained convolutional neural network (CNN) widely recognized for its remarkable accuracy in image recognition [12]. For shallow features, we use a popular local texture descriptor known as binarized statistical image features (BSIF) [13].

### C. Multilinear subspace learning using TXQDA coupled with WCCN

The feature vector is constructed by extracted features from the VGG16 and BSIF. The vector's large dimensionality is its primary issue. As a result, the feature vector must be projected into a lower space that only contains discriminant data [14]. In our work, we utilize a new multilinear subspace learning and dimensionality reduction named TXQDA proposed by [9] to enhance kinship verification [15], [9]. For kinship verification datasets, this novel technique produced superior outcomes to the earlier ones in terms of dimensionality reduction [9], [15]. The best multilinear projection matrices are estimated in the offline (training) stage, and new samples are projected by these tensors and matched in the online (test) stage. The histograms of several local descriptors taken from the training face images were used to generate the $X$, $Y$ $I_1 \times I_2 \times I_3$ training 3rd order tensors. The following definitions apply to the three modes of the tensors $X$ and $Y$ : $I_1$ relates to the local descriptors extracted at different scales, $I_2$ to the histograms, and $I_3$ to the dataset's face samples. Based on the TXQDA approach, the input tensors X and Z are reduced into $I_1$ and $I_2$ modes and projected into a different subspace. We then produce a reduced tensor where $i_1 \times i_2 \prec \prec I_1 \times I_2$. For more mathematical details about TXQDA, refer to [9], [15]. The Within-Class Covariance Normalization (WCCN) technique has been used to improve the speaker recognition system [16]. This method is used as an additional session variability compensation technique to scale the subspace features and to reduce the dimension of high within-class variance; as a result, the anticipated classification error in training data is decreased. Here, we suggest a TXQDA variation that integrates WCCN.

### D. Matching and logistic regression fusion

After projecting the facial images data by TXQDA algorithm, the matching process is performed computing using the cosine distance [16] between two vectors, $V_{t_1}$ and $V_{t_2}$, is defined by the the following equation.

$$\cos(V_{t_1}, V_{t_2}) = \frac{V_{t_1}^T . V_{t_2}}{\|V_{t_1}\| . \|V_{t_2}\|} \quad (1)$$

Matching refers to the process of comparing the extracted features or representations from two facial images and computing a similarity or dissimilarity score that quantifies their kinship relationship. To integrate the scores obtained from both deep and shallow texture features in our work, we utilized a powerful technique known as Logistic Regression (LR) [17]. Which has shown substantial performance enhancements in prior fusion studies [5], [18]. It makes a simple decision using a hyperplane and converges around an optimal solution. Operating the scores $Y_i$ as the input scores, the generated probability output by the LR process $X_i$ is defined as follows:

$$X_i = (1 + exp(aY_i + b))^{-1} \quad (2)$$

where $a$ represent a scalar factor and $b$ is a bias.

## III. EXPERIMENTS

In this section, we conduct a series of experiments to assess the performance of the proposed kinship verification system. We evaluate the system using three distinct datasets: Cornell Kin Face, UB Kin Face dataset and the TS Kin Face dataset. The experimental results obtained from these datasets are presented in Tables I, II and III respectively.

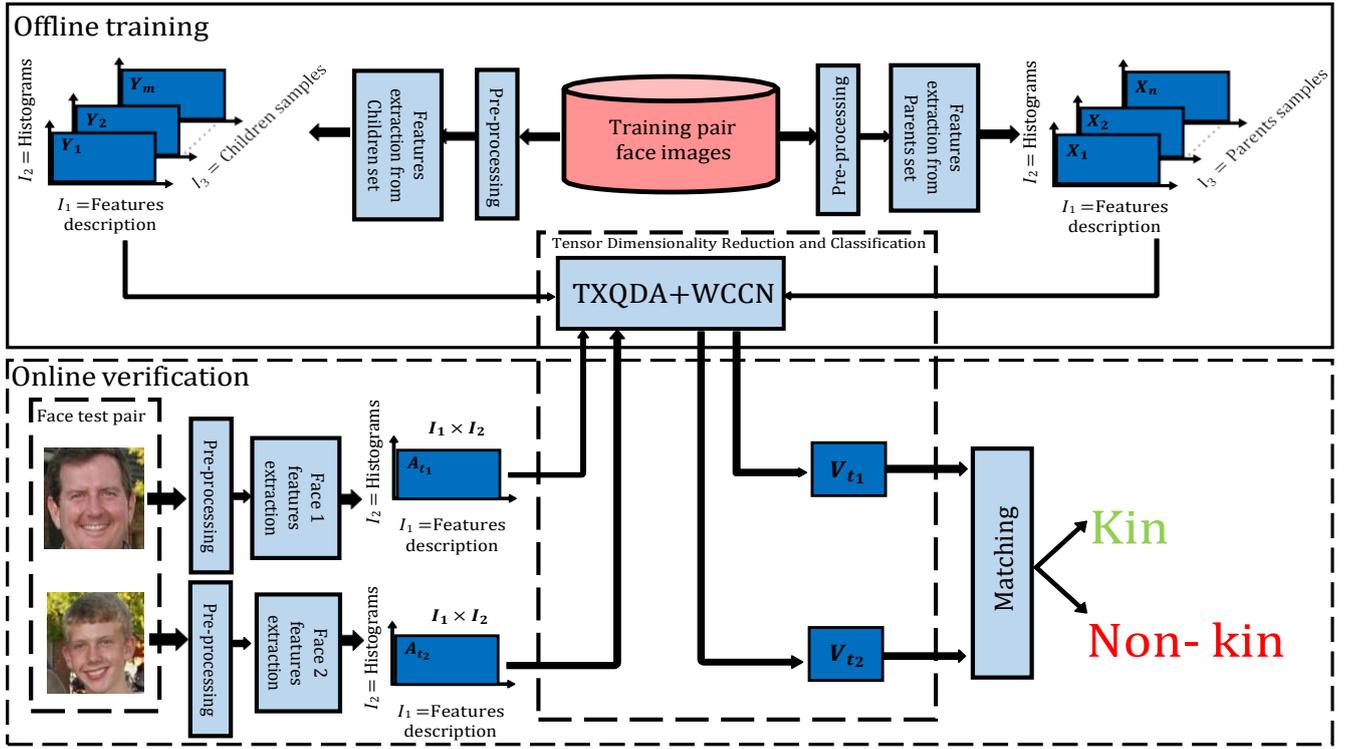

Fig. 1: Pipeline diagram of the proposed kinship verification and feature extraction using BSIF and VGG16

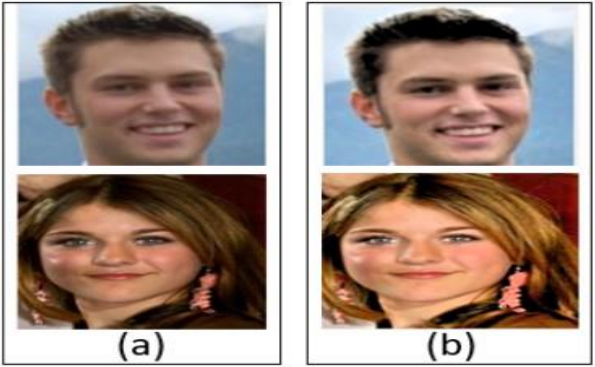

Fig. 2: Example of test images processing :(a) input images;(b) MSR images.

### A. Benchmark Datasets

Cornell Kinship [19]: The dataset being referred to consists of a total of 286 facial images, specifically related to 143 pairs of subjects. The facial images in this dataset depict subjects with a frontal pose and a neutral expression.

UB Kin Face [20]: This dataset consists of 600 images of 400 people, divided into 200 groups of child-young parent (set1) and 200 groups of child-old parent (set2). this dataset is regarded as the first of its kind, presenting a novel approach to the kinship verification problem, as it includes both young and old face images of parents.

TS Kin Face [21]: The Tri-subject kinship face dataset consists of images belonging to the child, mother, and father. The dataset comprises 513 images in the Father, Mother, and Son group, as well as 502 images in the Father, Mother, and Daughter group.

### B. Parameter Settings

In our experiments, we utilize the 5-fold cross-validation protocol [21], [22] to evaluate the performance of our approach. Prior to analysis, all face images in the datasets undergo pre-processing, specifically the detection of the facial region using the MTCNN method. Additionally, we employ MSR techniques to enhance the quality of the images. Subsequently, we extract two distinct types of features: Shallow Texture feature and Deep feature. For shallow texture features (BSIF), we use 6 flters with diferent sizes W = 3, 5, 7, 9, 11 and 13. The facial image is subsequently partitioned into 16 blocks. Each block is summed to create a histogram consisting of 256 bins. These individual histograms are then concatenated, resulting in a final feature vector with a size of $(1 \times 4096)$. For deep features(VGG16), we extract them from the face image with a size of $224 \times 224 \times 3$. We utilize two layers from the VGG16 network, specifically fc6 and fc7. The resulting feature from these layers are concatenated to form a feature matrix with a size of $(2 \times 4096)$.

### C. Result analysis and discussion

The experimental results of our study on the Cornell Kin Face and TS Kin Face datasets are presented in Tables 1 and 2. Tables I and III showcase the mean accuracy of the BSIF descriptor across its 6 scales W=3, 5, 7, 9, 11 and 13, as well as the fc6 and fc7 layers of the pretrained VGG16 model. Furthermore, these tables include the average accuracy resulting from fusing the scores of the best-performing BSIF and VGG16 results using LR fusion. Fig. 2 and 3 illustrates the ROC curves of the best results achieved by our approach.

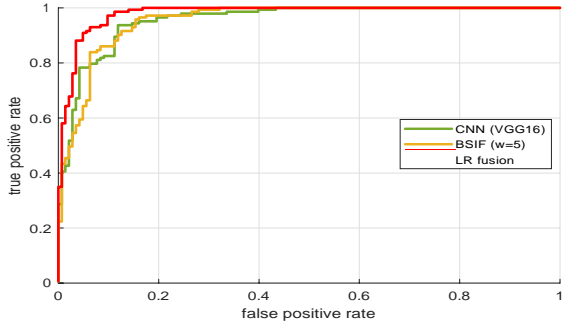

Fig. 3: ROC curves of the Shallow texture and Deep features on cornell datasets

TABLE I: The mean accuracy using BSIF on Cornell Kin Face dataset with different scales and the LR fusion with VGG16

| Features | Scales | Mean Acc (%) |
|---|---|---|
| BSIF without MSR | w=5 | 85.77 |
| BSIF with MSR | w=3 | 91.10 |
|  | w=5 | 92.01 |
|  | w=7 | 91.43 |
|  | w=9 | 90.16 |
|  | w=11 | 90.09 |
|  | w=13 | 89.78 |
| VGG16 | fc6 and fc7 | 91.02 |
| LR Fusion | BSIF (w=3) + VGG16 | **94.82** |

TABLE II: The mean accuracy using BSIF on UB Kin Face dataset with different scales and the LR fusion with VGG16.

| Features | Scales | set1 | set2 | Mean Acc (%) |
|---|---|---|---|---|
| BSIF without MSR | w=5 | 75.19 | 74.34 | 74.76 |
| BSIF with MSR | w=3 | 81.44 | 79.88 | 80.66 |
|  | w=5 | 85.42 | 85.13 | 85.27 |
|  | w=7 | 83.76 | 82.90 | 83.33 |
|  | w=9 | 85.71 | 82.24 | 83.97 |
|  | w=11 | 84.99 | 83.46 | 84.22 |
|  | w=13 | 84.45 | 83.40 | 83.92 |
| VGG16 | fc6 and fc7 | 87.28 | 85.21 | 86.24 |
| LR Fusion | BSIF (W=3) + VGG16 | **92.13** | **91.75** | **91.94** |

*D. Discussion*

Based on the experiments conducted using the proposed approach, which depends on fusion using two kinds of features (Deep and Shallow texture), on the three datasets (Cornell Kin Face, UB Kin Face and TS Kin Face), we have reached the following conclusions: The influence of the image histogram is significant in enhancing the precision of metric evaluation. Additionally, we took advantage of applying the preprocessing technique based on MSR. The obtained results clearly demonstrate the powerful performance of LR fusion when compared to using each type of feature independently. In our study, we employed score-level fusion using two scores generated by the CNN-based VGG16 with the BSIF descriptor. Through the LR fusion process, we achieved high performance with accuracy rates of 94.82 % on the Cornell Kin Face dataset, 91.94 % on UB Kin Face and 90.77 % on the TS Kin Face dataset, as shown in Tables I and III respectively.

TABLE III: The mean accuracy using BSIF on TS Kin Face dataset with different scales and the LR fusion with VGG16.

| Features | Scales | FS | FD | MS | MD | Mean Acc (%) |
|---|---|---|---|---|---|---|
| BSIF without MSR | w=5 | 81.13 | 80.03 | 79.98 | 78.56 | 79.92 |
| BSIF with MSR | w=3 | 85.95 | 86.94 | 87.24 | 88.23 | 87.12 |
|  | w=5 | 88.61 | 87.13 | 86.83 | 86.92 | 87.37 |
|  | w=7 | 86.83 | 86.24 | 86.83 | 86.53 | 86.60 |
|  | w=9 | 87.43 | 86.03 | 87.91 | 86.52 | 86.97 |
|  | w=11 | 85.14 | 85.01 | 87.97 | 86.26 | 86.09 |
|  | w=13 | 85.30 | 82.28 | 83.50 | 84.44 | 83.70 |
| VGG16 | fc6 and fc7 | 77.38 | 78.12 | 79.32 | 79.70 | 78.63 |
| LR Fusion | BSIF (W=3) + VGG16 | **90.99** | **89.21** | **90.30** | **92.57** | **90.77** |

TABLE IV: Performance comparison of kinship verification state of the art on Cornell Kin Face and TS Kin Face datasets

| Work | year | Cornell dataset | UB dataset | TS dataset |
|---|---|---|---|---|
| MSIDA [8] | 2019 | 86.87 | 83.34 | 85.18 |
| SP-DTCWT [23] | 2020 | - | 72.87 | - |
| FMRE2 [24] | 2021 | 84.16 | 85.03 | 90.85 |
| RDFSA [25] | 2021 | - | 79.81 | 85.02 |
| AdvKin [26] | 2021 | 81.40 | 75.00 | - |
| BC2DA [27] | 2022 | 83.07 | 83.30 | 83.55 |
| TXQEDA [28] | 2022 | 93.77 | - | 90.68 |
| **Proposed** | **2023** | **94.82** | **91.94** | **90.77** |

*E. Comparison against the state of the art*

The effectiveness of our proposed method, which involves fusing BSIF and VGG16 scores using the LR fusing technique, is compared to more modern methods in Table IV for the Cornell Kin Face, UB Kin Face and TS Kin Face datasets. The comparison clearly shows that our proposed technique outperforms the recent state-of-the-art methods on the both datasets Cornell Kin Face, UB Kin Face and TS Kin Face.

## IV. CONCLUSION

The work presents a kinship verification system that utilizes a novel and efficient approach for face description by employing LR fusion between deep (VGG16) and shallow texture (BSIF) features. Leveraging tensor subspace learning, our approach demonstrates impressive performance. Our findings indicate that deep and handcrafted texture characteristics mutually complement each other at the score level, where the fusion significantly improves kinship verification accuracy.

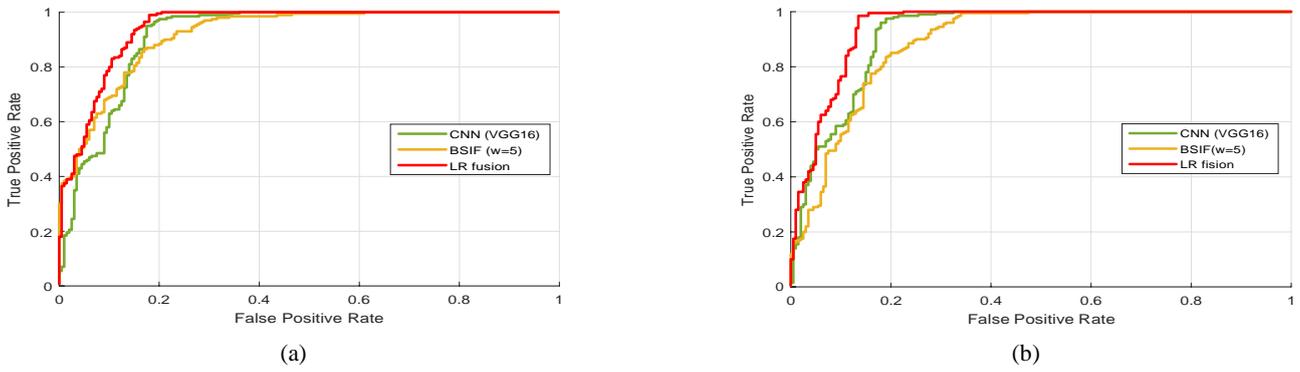

Fig. 4: ROC curves of the Shallow texture and Deep features on UB datasets, (a) set1 and (b) set2.

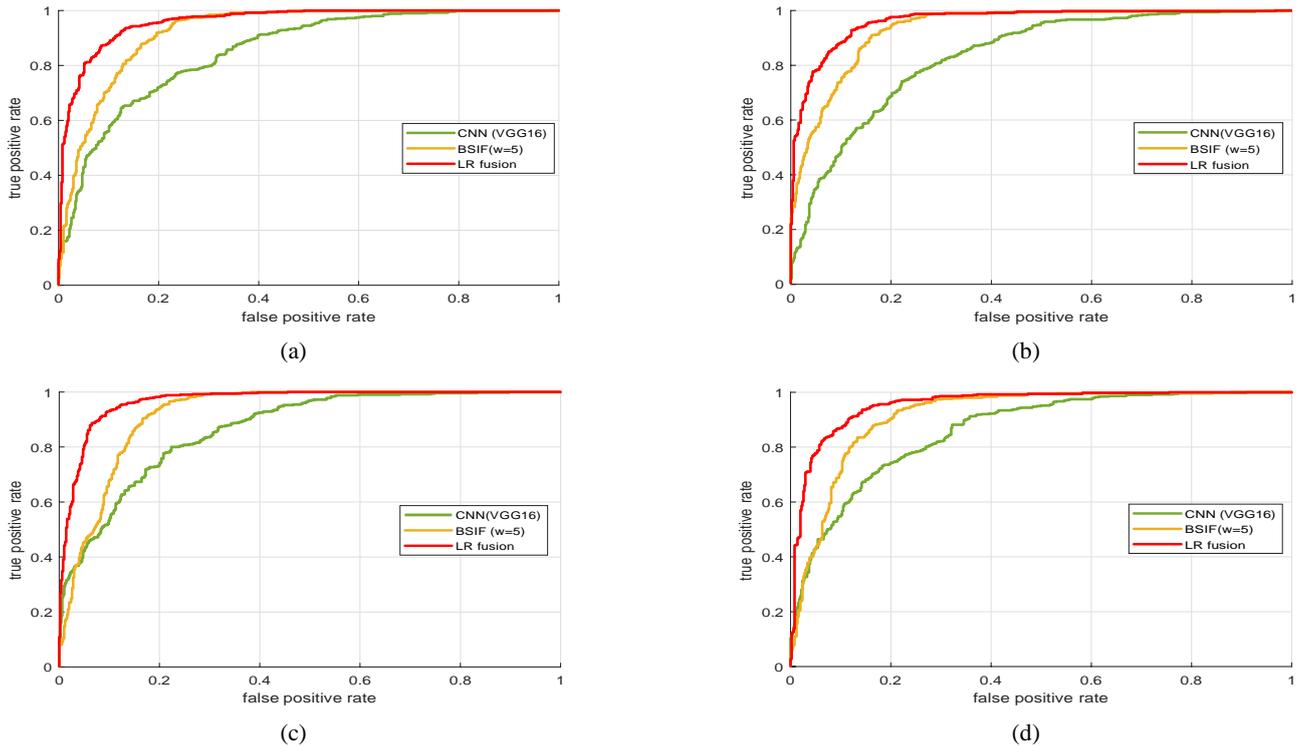

Fig. 5: ROC curves of the Shallow texture and Deep features on TS datasets, (a) F–D set, (b) F–S set, (c) M–D set and (d) M–S.